\documentclass{article}
\usepackage{graphicx}
\usepackage{amsmath}
\usepackage{hyperref}
\usepackage{enumitem}
\usepackage{longtable}
\usepackage{booktabs}
\usepackage{multirow}
\usepackage{amssymb}
\usepackage{float} 
\usepackage{placeins} 

\title{Performance of Large Language Models in Numerical vs. Semantic Medical Knowledge: Benchmarking on Evidence-Based Q\&As}
\author{
    Eden Avnat$^{1,2}$, Michal Levy$^{2,3,4}$, Daniel Herstain$^{1,2,5}$, Elia Yanko$^{6}$, \\
    Daniel Ben Joya$^{2,7}$, Michal Tzuchman Katz$^{2}$, Dafna Eshel$^{2}$, \\
    Sahar Laros$^{1,2}$, Yael Dagan$^{1,2}$, Shahar Barami$^{1,2}$, \\
    Joseph Mermelstein$^{2}$, Shahar Ovadia$^{2}$, Noam Shomron$^{1}$, \\
    Varda Shalev$^{1}$, Raja-Elie E. Abdulnour$^{8}$
}
\date{}

\begin{document}
\maketitle

\begin{center}
    $^{1}$Faculty of Medicine, Tel-Aviv University, Tel-Aviv, IL \\
    $^{2}$Kahun Medical Ltd, Givatayim, Israel \\
    $^{3}$Faculty of Medicine, Hebrew University of Jerusalem, Jerusalem, IL \\
    $^{4}$School of Computer Science and Engineering, The Hebrew University of Jerusalem, Jerusalem, IL \\
    $^{5}$Leumit Health Services, Tel Aviv-Yafo, IL \\
    $^{6}$The Azrieli Faculty of Medicine, Bar-Ilan University, Safed, IL \\
    $^{7}$Kaplan medical center, Rehovot, IL \\
    $^{8}$Brigham and Women’s Hospital, Harvard Medical School, Boston, MA \\
\end{center}

\begin{abstract}
    \textbf{Objective:} Clinical problem-solving requires processing of semantic medical knowledge such as illness scripts and numerical medical knowledge of diagnostic tests for evidence-based decision-making. As large language models (LLMs) show promising results in many aspects of language-based clinical practice, their ability to generate non-language evidence-based answers to clinical questions is inherently limited by tokenization. Therefore, we evaluated LLMs' performance on two question types: numeric (correlating findings) and semantic (differentiating entities) while examining differences within and between LLMs in medical aspects and comparing their performance to humans.

    \textbf{Methods:} To generate straightforward multi-choice questions and answers (QAs) based on evidence-based medicine (EBM), we used a comprehensive medical knowledge graph (encompassed data from more than 50,00 peer-reviewed articles) and created the “EBMQA”. EBMQA contains 105,222 QAs labeled with medical and non-medical topics and classified into numerical or semantic questions. We benchmarked this dataset on two state-of-the-art LLMs, Chat-GPT4 and Claude3-Opus. We evaluated the LLMs accuracy on semantic and numerical question types and according to sub-labeled topics. For validation, six medical experts were tested on 100 numerical EBMQA questions.

    \textbf{Results:} In an analysis of 24,542 QAs, Claude3 and GPT4 performed better on semantic QAs (68.7\% and 68.4\%, respectively) than on numerical QAs (63.7\% and 56.7\%, respectively), with Claude3 outperforming GPT4 in numeric accuracy (p$<$.001). A median accuracy gap of 7\% [5-10] was observed between the best and worst sub-labels per topic, with different LLMs excelling in different sub-labels. Furthermore, humans (82.3\%) surpassed both Claude3 (64.3\%, p$<$.001) and GPT4 (55.8\%, p$<$.001) in the validation test.

    \textbf{Conclusions:} Both LLMs excelled more in semantic than numerical QAs, with Claude3 surpassing GPT4 in numerical QAs. However, both LLMs showed inter and intra gaps in different medical aspects and remained inferior to humans. Thus, their medical advice should be addressed carefully.

\end{abstract}

\section{Introduction}
Clinical problem-solving requires the processing of data using the clinician’s fund of knowledge in the form of illness scripts \cite{1,2}, most of which is semantic. The statistical weight of relationships between data that define an illness is the numerical equivalent of medical knowledge that is essential for prioritizing diagnostic hypotheses and decision-making \cite{3}.

Clinicians develop and use numerical knowledge through original research, and leverage diagnostic support tools for more complex decision-making \cite{4,5}. However, the explosive amount of medical knowledge and complex healthcare systems are tremendous challenges to high-quality, evidence-based medicine (EBM) \cite{6,7}.

The breakthrough of Large Language Models (LLMs), which process extensive data and encode knowledge from numerous online studies, shows great promise as tools for medical decision support \cite{8,9}. LLMs provide users with a sense of reliability and accuracy, but evidence shows that they occasionally generate responses that are not based on actual knowledge or give incorrect explanations \cite{10,11}. In addition, their performance on non-textual knowledge like medical codes is limited \cite{12}. Thus, physicians continue to express skepticism regarding LLMs and their capacity to outperform humans \cite{13}.

Several benchmark studies have addressed this subject by focusing on lengthy questions from licensing exams \cite{8,14} or on datasets derived from medical abstracts that could only be answered with yes/no/maybe \cite{15}.

To create a dataset that consists solely of EBM knowledge and is flexible enough to generate both semantic and numeric QA we used the Kahun knowledge graph- a clinically validated artificial intelligence tool that uses a medical, evidence-based knowledge graph. We have developed a methodology to generate QAs from this knowledge graph and created the EBMQA dataset. The dataset comprises 105,222 short multiple-choice questions, based on insights extracted from full-length articles, and aimed to test LLMs ability to assist physicians.

Finally, we benchmarked two state-of-the-art LLMs: Open-AI’s Chat-GPT4 (GPT4) \cite{16}, and Antropic’s Claude3 Opus (Claude3) \cite{17}, using part of EBMQA. Additionally, we compared their results to medical experts. Thus, we could evaluate the performance of LLMs in both numerical and semantic QA, identify differences within and between LLMs across diverse medical and non-medical domains, and compare their results to humans.

These analyses allowed us to assess whether physicians can trust LLMs.
\section{Methods}

\subsection{EBMQA}

\subsubsection{Kahun}
Kahun is a diagnostic tool based on artificial intelligence and structured knowledge graph technologies. The knowledge graph encompasses more than 50,000 peer-reviewed publications and more than 20,000,000 medical relations that were mapped by medical experts \cite{18}. Kahun’s unique structure and its EBM content serves as a reasonable platform to generate the EBMQA.

\subsubsection{Questions Structure}
All QAs were derived from Kahun’s knowledge graph. Each question was generated based on data from nodes and edges in the graph and consisted of three main entities: source (usually a disorder that is related to the target), target (usually a symptom or sign that is related to the source), and background (usually a relevant population related to the source). In this study, we refer to source, target, or background as entities. Additionally, the relation between entities (derived from data on the edges) determines the question type and the specific template that was used to generate the question and the answers.Further explanation regarding template creation provided in the Appendix.

EBMQA is comprised of two types of questions:
\begin{enumerate}
    \item Numeric QAs - derived from connections between a source and a single target. Those questions deal with choosing the range in which the correct answer resides (Figure 1).
    \item Semantics QAs - derived from connections between a source and up to six targets (possible answers). Each QA deals with choosing the most common target/s related to a source given a specific relation (subtype, location, duration, etc) (Figure 1).
\end{enumerate}
Further examples of both numeric and semantics QAs are in Table S1.

\begin{figure}[H] 
    \centering
    \includegraphics[width=1\textwidth]{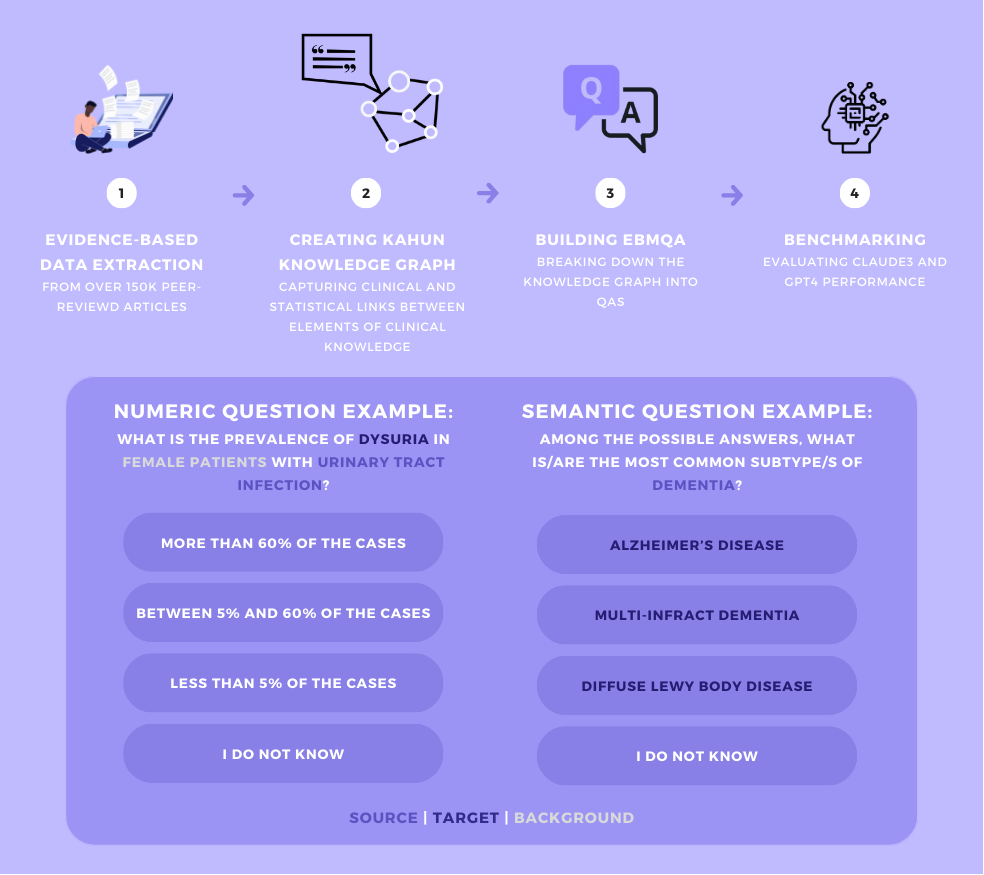}
    \caption{The flowchart of the study: From Kahun's knowledge graph, which references source, target, and background as edges of the graph (1-2), to the EBMQA dataset and the LLM benchmarking (3-4), which includes both numeric and semantic QAs.}
    \label{fig:study_flowchart}
\end{figure}
\FloatBarrier 

\subsubsection{Multiple Choice Questions Structure}
The questions in the EBMQA are multiple-choice questions. Numeric QAs have one correct answer, while semantic QAs have up to five correct answers. However, if one does not know what the answer to the question is, an “I do not know” (IDK) option was added to all QAs as a possible answer.

\subsubsection{Numerical Data and Possible Answers}
Each QA is based on numerical data derived from Kahun's knowledge graph. This includes minimum, maximum, and mid values estimating the connections between medical entities. We employed statistical methods, including median and median absolute deviation (MAD), to categorize answers into meaningful ranges based on their calculated mid-values. Specific methodologies for categorizing these ranges, and detailed statistical backing for each type of QA, are documented in the Appendix.

\subsubsection{QA exclusion}
EBMQA aims to provide simple short medical QA, therefore, questions related to multiple sources, backgrounds, or targets (except semantic questions) were excluded from the EBMQA. Additionally, QAs that are not related to medical knowledge (such as the average length of a season), were removed. Duplicated questions were excluded from the EBMQA, though the answer for the remaining question is the average of all duplicated Mean values. QAs with all or no correct answers were deleted from the dataset.

\subsubsection{Labeling }
Each QA in the study was categorized using multiple medical data labels which were derived from standardized medical classifications such as those provided by Snomed CT [19] and Kahun's medical expertise. These classifications include, but are not limited to, Medical Type, Medical Subject Type, Medical Discipline, and Prevalence. Each QA was also analyzed for its question length and distribution of answers. Details on the specific labeling criteria and categories are described in the Appendix.

\subsection{Benchmark Analysis}
\subsubsection{QA selection and subanalysis}
Due to the relatively different structure of semantic QA and its small number of QA, we analyzed all of them separately. 
Regarding numeric QA types and in the search for meaningful parameters that might influence LLM’s performance, the benchmark included QAs based on three medical (Medical subject type, Medical Discipline and Prevalence) and three non-medical sub-labels (QA types, Question length and Answers distribution) as further detailed in Appendix (Figure S1). All QAs were selected randomly, and although the total number of QAs per label varied, each label contains an identical number of selected QAs per sub-labeled entity. Additionally, each QA was selected only once.

\subsubsection{LLMs prompting}
In this study, we used two state-of-the-art LLMs: Chat-GPT4 (gpt-4-0125-preview), and Claude3-Opus (claude-3-opus-20240229). Both models’ parameters were set to be: temperature= 0 and maximum tokens= 300. All queries were sent to the LLMs using R-studio (version 4.2.2). Further descriptions of the prompts and suitable examples are documented in the Appendix.

\subsubsection{Evaluating LLM’s performance}
We evaluated LLM’s performance using the following metrics:
\begin{enumerate}[label=\textbullet]
    \item Accuracy - for both semantic and numeric QAs, the total number of correct answers suggested by the LLM divided by the total answers suggested by the LLM, excluding IDK answers.
    \item Answer rate (AR) - for both semantic and numeric QAs, the total number of both correct and wrong answers suggested by the LLM divided by the total answers suggested by the LLM, including IDK answers.
    \item Majority - the distribution of the most frequent correct answers among the total correct answers.
\end{enumerate}

\subsubsection{Prompt sensitivity analysis}
To test both the effects of adding IDK as a possible answer and changing the order of answers (including IDK), eight different prompts were tested on 100 numerical questions that were randomly selected (questionnaire): four prompts, each with a different order of possible answers, including IDK, and four prompts, each with different order of possible answers, without IDK.

\subsubsection{Human validation}
To validate the QAs in the EBMQA, two physicians and four medical students (in their clinical years) answered the questionnaire, first with the IDK possibility and then with guessing the answer to their IDK questions. Their accuracies with and without guessing were compared to LLMs'. 

\subsubsection{Analysis and variables}
All statistical analyses were performed using R-studio (R version 4.2.2). Categorical variables were represented by a percentage while continuous variables were represented by mean and standard deviation (SD) if distributed normally or by median and interquartile range (IQR) else. The cutoff for statistically significant results was set at was set at alpha= 0.05 and confidence intervals (CI) were set to 95\%. Proportions comparison was conducted using the "Proportion test". For estimating the correlation of two quantitative variables, we employed the Spearman correlation. 

\subsubsection{Ethics}
This study was approved by the Tel Aviv University Ethics Committee (Institutional Review Board (IRB) protocol number 0008527-2).

\section{Results}

\subsection{EBMQA}
The EBMQA contains 105,222 QAs. Additionally, each QA pair was labeled according to metadata labels and medical labels.

\subsubsection{Medical labels}
The EBMQA encompasses diverse medical data types, including a unique count of 7,746 “Disorders”, 2,547 “Signs or Symptoms”, 1,243 “Lab tests”, 885 “Imaging or procedures” data, 474 “Background” data (demographics, habits, family history, etc.), and more (Figure S2A).

Among the Medical subject types, “Disorders” was the most abundant with 45,964 QAs, followed by “Symptoms and Signs” with 30,152 QAs, “Lab test” with 5,966 QAs, and “Imaging or Procedures” with 4,374 QAs. All the other subjects encompass 640 QAs (Figure S2B).

Focusing on Medical Discipline, the EBMQA contains 64,846 relevant QAs: the leading medical discipline was the Digestive system with 9,879 QAs, followed by Cardiovascular system with 7,847 QAs and Infectious diseases with 7,798 QAs. Musculoskeletal system had the least number of QAs- 2,832 (Figure S2C).

Regarding the “Prevalence” label, the median prevalence was 1e-4 and MAD of 9.810102e-05. Among them 36,653 QAs focused on high-prevalence disorders, 22,139 QAs focused on moderate-prevalence disorders, and 2,531 QAs focused on low-prevalence disorders. (Figure S2D)

\subsubsection{Metadata labels}

EBMQA includes 13 distinct QA types (Figure S2E). The most frequent QA type is “Sensitivity” with 70\% (74,140/105,222) of the total QAs, while eight QAs have less than 900 QAs per QA type: Specificity, PLR (positive likelihood ratio), NLR (negative likelihood ratio), Relative Risk, Prevalence, PPV (positive predictive value), NPV (negative predictive value), and Associated Risk.

In total, the median number of words per question (including the question, the instructions and the possible answers) in the EBMQA was 57 [IQR: 53-66] with a MAD of 5. The medium question length group has the majority of QAs, with 59,998 QAs, whereas the short question length group, with 9,968 QAs, represents the minority (Figure S2F). Focusing on each QA type reveals that the “Risk Factor” QA type has the longer median of words per question with 81 [IQR: 80-84] words while the “Sensitivity” QA type has the shorter median words per question with 54 [IQR: 52-58] words (Table S2). 

Regarding numeric questions with three range values, the most frequently distributed answer was the mid-range values (46,431 QAs), followed by the low-range values (28,598 QAs) and the high-range values (14,292 QAs) (Figure S3).

\subsection{Benchmark Analysis}
Out of the 105,222 QA in the EBMQA, the same 24,542 questions were presented to each LLMs. The "numeric" QA analysis comprised 90\% (22,000/24,542) of the questions, whereas "semantic" QA analysis accounted for the remaining 10\% (2,542/24,254).

Both LLMs demonstrated better performances in the semantic QAs than in the numeric QAs in terms of accuracy and AR (Calude3- 68.65\% (1,592.78/2,320) vs 61.29 (8,583/14,005), p$<$.001, GPT4- 68.38\% (1708.85/2499) vs 56.74\%
( 12,038/ 21,215), p$<$.001 and Calude3- 94.62\% (2,320/2,542) vs 63.66\% (14,005/ 22,000), p$<$.001, GPT4- 98.31\% (2,499/2,542) vs 96.4\% (21,215/22,000), p$<$.001 respectively).
From an inter-model perspective, Calude3 outperformed GPT4 in numeric accuracy though no significant difference was found in semantic accuracy. However, in comparison to Calude3, GPT4 had a higher AR in both semantic and numeric questions (Table1).

\begin{table}[h]
\centering
\resizebox{\textwidth}{!}{%
\begin{tabular}{lcccccc}
\toprule
\textbf{Question Type} & \textbf{Metric} & \textbf{Claude3} & \textbf{GPT4} & \textbf{P-value (Chi-squared)} & \textbf{OR} & \textbf{CI} \\
\midrule
\multirow{2}{*}{Semantic} & Accuracy & 68.65\% & 68.38\% & 0.8364 & 1.01 & [0.9-1.14] \\
 & (count) & (1,592.78/2,320) & (1,708.85/2,499) &  &  &  \\
 & Answer-rate & 94.62\% & \textbf{98.31\%} & $<$.00001 & 0.18 & [0.13-0.25] \\
 & (count) & (2,320/2,542) & (2,499/2,542) &  &  &  \\
\midrule
\multirow{2}{*}{Numeric} & Accuracy & \textbf{61.29\%} & 56.74\% & $<$.00001 & 1.2 & [1.16-1.26] \\
 & (count) & (8,583/14,005) & (12,038/21,215) &  &  &  \\
 & Answer-rate & 63.66\% & \textbf{96.4\%} & $<$.00001 & 0.06 & [0.06-0.07] \\
 & (count) & (14,005/22,000) & (21,215/22,000) &  &  &  \\
\bottomrule
\end{tabular}%
}
\caption{Comparison of Claude3 and GPT4 on Semantic and Numeric Questions}
\end{table}

\subsubsection{Prompt sensitivity analysis}
The average accuracy of Claude3 with the IDK option vs without it was not significantly different ($64.25\% \pm 3.95$  vs  $59.25 \pm 5$, $p = .17$). The same trend was noted in GPT-4 ($55.75\% \pm 1.71$ vs $53.25\% \pm2.89$, $p = .24$). Additionally, within each subgroup: Claude3 with the IDK option, Claude3 without the IDK option, GPT4 with the IDK option, and GPT4 without the IDK option- no single answer-option-order prompt was significantly superior to the others (Figure S4, Figure S5).

\subsubsection{Human Validation}

Calude3 and GPT4 average accuracy rates, with the IDK option or without it were higher than random (33\%) or majority guessing (47\%) though had lower average accuracy rates in comparing to humans with the IDK option ($82.3\% \pm2.82$) or without it ($78.2\% \pm3.6$) (Figure 2, Table S3).

\begin{figure}[H] 
    \centering
    \includegraphics[width=1\textwidth]{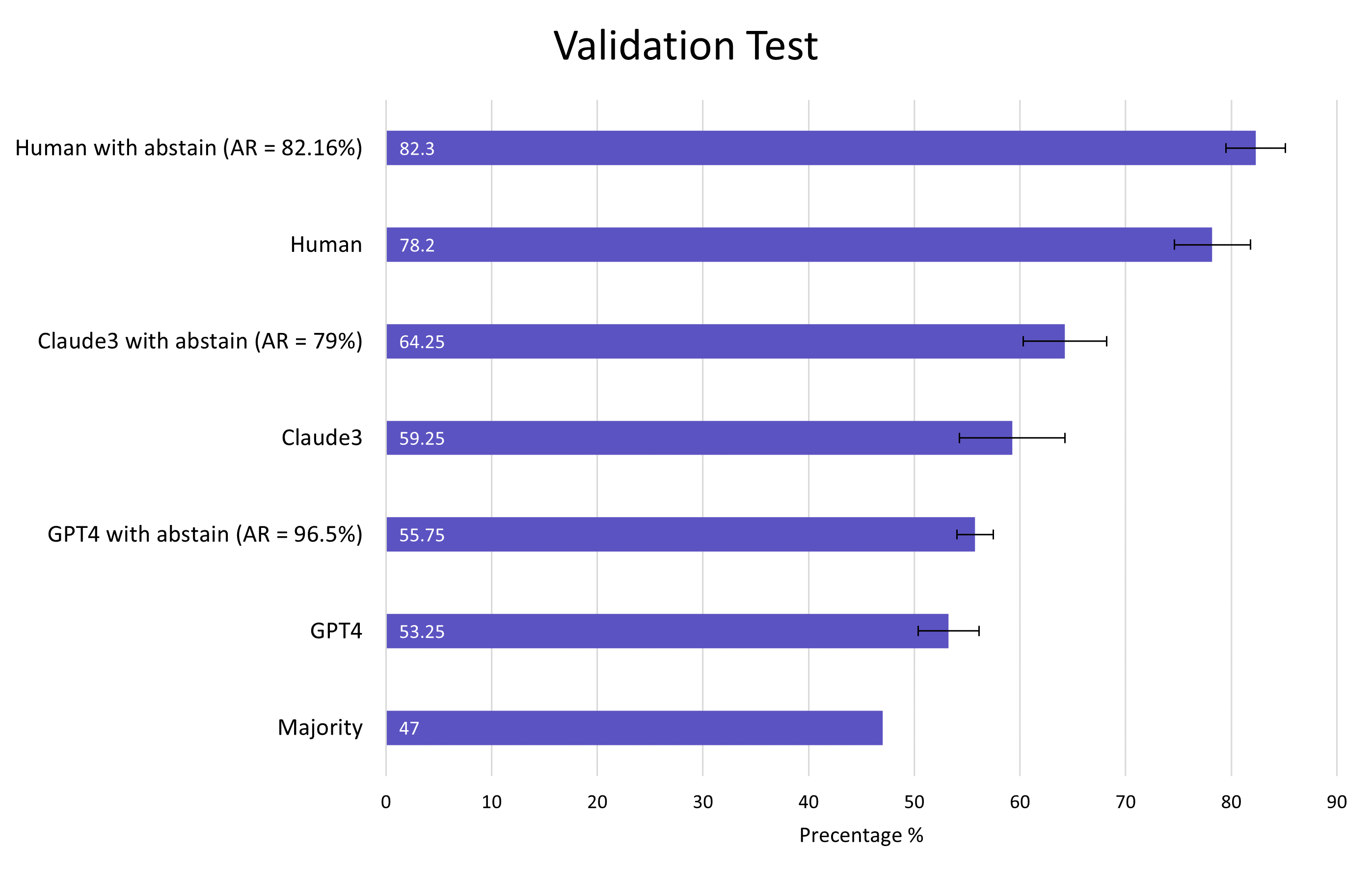}
    \caption{Validation test: Each LLM was tested eight times- four times with the option to “I do not know” (abstain), using the same prompt though in a different order of possible answers, and four times without the abstain option, using the same prompt though in a different order of possible answers. Additionally, six medical experts were tested: first, with the option to abstain, and then without. Confidence intervals of 95\% were calculated accordingly while answer-rate (AR) were added only to abstaining instances.}
    \label{fig:study_flowchart}
\end{figure}
\FloatBarrier 

\subsubsection{Numeric QA subanalysis}

The accuracy gap between the highest and lowest accuracy rates for each label in each LLM was calculated, revealing a median difference of 7\% [IQR: $5\%-10\%$] (Figure 3). Focusing on disorders selected sub-labels, Claude3 performed well in neoplastic disorders but struggled with genitourinary disorders (69\% (676/984) vs 58\% (464/803), p$<$.0001), while GPT4 excelled in cardiovascular disorders but struggled with neoplastic disorders (60\% (1076/1783) vs 53\% (704/1316), $p=.0002$, Table S4). Furthermore, among sub-labels disorders queried over 200 times, Spearman’s correlations between question-answer rate and accuracy rate in both Claude3 and GPT4 was insignificant ($\rho=0.12$, $\rho=.69$; $\rho=0.43$, $\rho=.13$).

\begin{figure}[H] 
    \centering
    \includegraphics[width=1\textwidth]{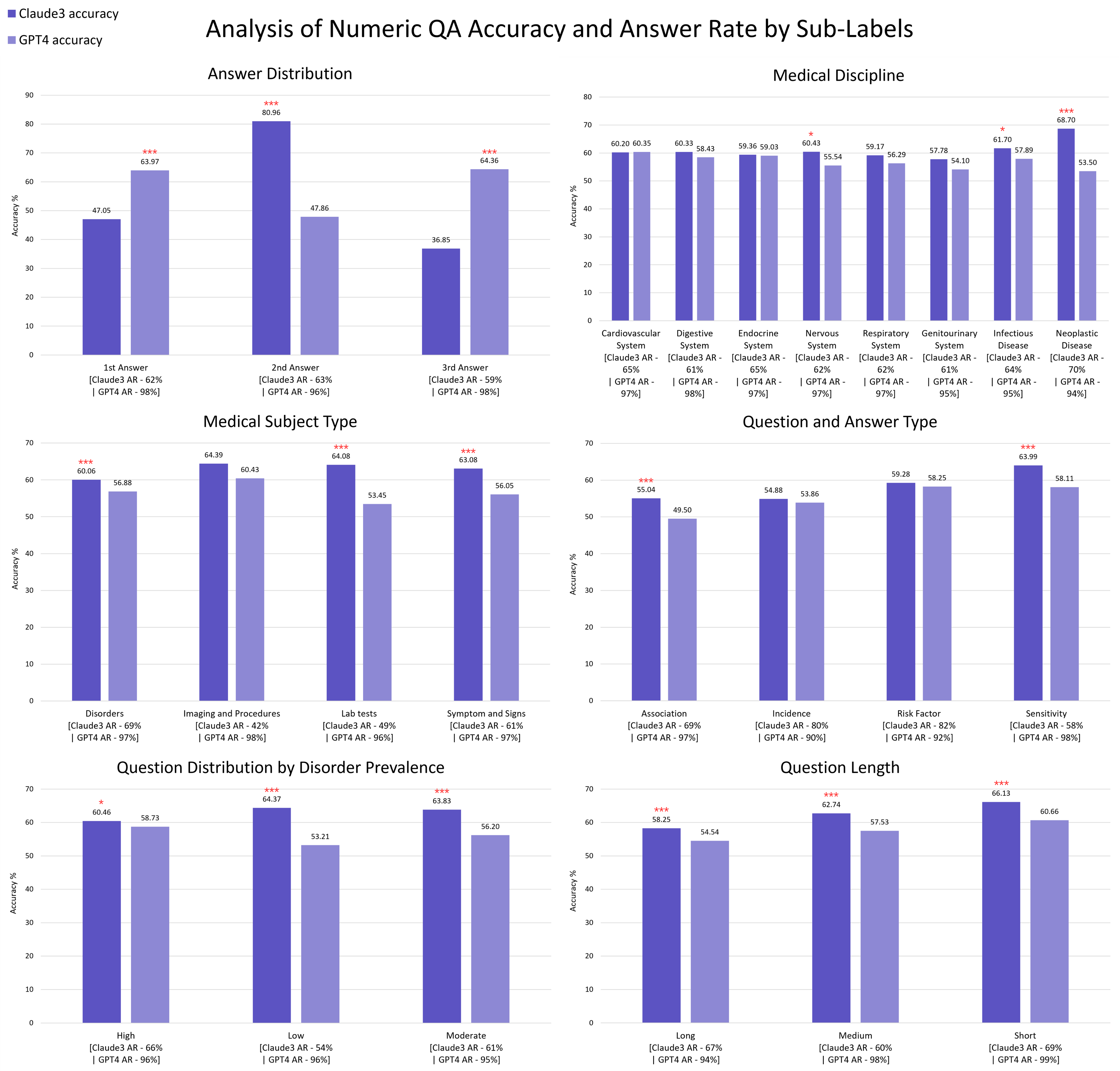}
    \caption{Numeric QA accuracy and answer-rate sub-labels analysis: (A) Answer distribution, (B) Medical Discipline, (C) Medical Subject type, (D) QA type, (E) Disorders Prevalence, (F) Question length.Red asterisks represent proportion p-values: .05$<$ *$<$.01, ***$<$.0001}
    \label{fig:study_flowchart}
\end{figure}
\FloatBarrier 

\section{Discussion}

This study aimed to highlight the current gaps in the medical knowledge of LLMs and their current ability to surpass humans. We presented a method to create a QA dataset (EBMQA) from a structured knowledge graph and benchmarked two state-of-the-art LLMs (GPT4 and Claude3) \cite {16,17}. We demonstrated that both LLMs performed better in semantic QAs than in numerical QAs by asking more than 24,000 QAs (Table1). Claude3 outperformed GPT4 in numerical QAs, and showed similar results in semantic QAs though had significantly lower ARs (Table1). A validation test indicated that the numerical accuracy rates of Claude3 and GPT4 were higher than a majority-guessing, though lower than medical experts (Figure 2). 

The use of knowledge graphs for evaluating LLMs is gaining popularity \cite{20,21,22}. Kahun’s structured knowledge graph enabled us to generate both semantic and numeric labeled QA, without utilizing advanced models \cite{22}. Our QAs generation process, which relies on templates designed to fit a source-target-background graph structure can be applicable to other graphs with a similar structure. Additionally, this relatively large knowledge graph allowed us to create a massive EBM dataset. Moreover, we embraced a data-driven approach in which distractors were based on sub-analysis distribution rather than specific or random values

The EBMQA, which consists of 105,222 straightforward single-line QAs, was designed to mimic physicians’ strategy of breaking complex medical scenarios into less complicated problems, unlike medical licensing examination datasets which are typically complex-cases oriented \cite{14,23}. Additionally, the EBMQA addresses numeric/semantic data which is considered fundamental for physicians \cite{24,25} while dealing with data from articles and embracing the EBM approach \cite{7}, as opposed to the abstracted-based yes/no/maybe QAs in PubMedQA \cite{15}.

A major concern regarding applying LLMs in healthcare is the uncertainty of providing solid evidence which supports their answers \cite{8}. Clinical evidence predominantly relies on statistical and numerical data. Thus, it is imperative to examine whether LLMs can deliver this type of reasoning. It has been shown that LLMs are more capable when given semantic questions rather than numerical questions though in a relatively low sample size (smaller than 200 QAs) \cite{26}. As far as we know, we were the first to show this trend in the medical field while using a much larger scale (Table1). Furthermore, since both semantic and numeric questions in the EBMQA may address the same entities but from different perspectives, it is questionable whether LLMs can support their semantic answers with statistical data.

A recent benchmark analysis, focused on nephrology QAs only, found that GPT4 outperformed Claude2 \cite{27}. Although our inter-model examination did not include a direct nephrology compression, due to a different classification method, it reveals that generally Claude3 outperformed GPT4 and specifically in a variety of medical disciplines such as neoplastic disorders, nervous system and more (Figure 3B). Those results raise the need to constantly benchmark new LLMs as they continuously improve.

Regarding internal model variations, the differences in accuracy between the highest and lowest performing medical disciplines- 8\% for Claude3 and 6\% for GPT4 (Figure 3B), support previous benchmarks which found that the performance of LLMs can vary across different medical disciplines \cite{27,28}.

Moreover, this comprehensive benchmark widens the medical scope and further supports both intra- and inter-model differences by exploring medical subjects: as Claude3 favors “Imaging and Procedures” and struggles with “Disorders” (64\% (463/719) vs 60\% (3181/5296), $p=.03$, respectively), GPT4 excel in “Imaging and Procedures” but struggles with “Lab tests” (60\% (1017/1683) vs 53\% (1008/1886), $p<.0001$, respectively). 

As the debate over whether models surpass humans persists \cite{27,28}, the outcomes of our validation tests suggest that humans still excel in certain medical tasks. Therefore, we support further evaluations of LLMs before using them in medical settings.

Furthermore, the insignificant correlation between accuracy and AR, contradicts the theory that a model's confidence in its response reflects its subject expertise \cite{29}. Thus, abstaining from providing an answer failed to explain the intra-model variance results, specifically in the medical discipline. Therefore, we are concerned that without previous knowledge regarding both the medical discipline and the model, the trustworthiness of LLMs is doubtful. Thus, we encourage further research on this subject.

In terms of prompt engineering, our sensitivity analysis showed relatively small SDs in prompt accuracy which supported our prompt stability. Additionally, although insignificant, the IDK prompt yielded higher accuracy and, therefore was used. Moreover, changing the order of the distractors did not significantly affect the performances of the LLMs.

Our benchmark did not include medical-tuned LLMs, which showed promising results \cite{30}, since they are not publicly available. Therefore, a similar benchmark including those LLMs is highly recommended. Moreover, we believe that nowadays, physicians are asking LLMs straightforward questions without adding further information or using external methods like Retrieval-Augmented Generation. Thus, those methods, that might improve the results, are not within the scope of this study, and further research is needed.  

\section{Conclusions}

On the EBMQA dataset, which resembles physicians’ problem-solving approach, LLMs were better at solving semantic than numeric questions. Despite Claude3 surpassing GPT4, both LLMs exhibited inter and intra gaps in medical knowledge. Additionally,  humans outperformed both LLMs in numeric questions. These results suggest that LLMs responses, especially numeric ones, should be considered cautiously in clinical settings.

\section{Abbreviation}
 
LLM- Large language model ; QA- Question and answer ; IQR- Interquartile range ; MAD- Median absolute deviation ; PLR- Positive likelihood ratio ; NLR- Negative likelihood ratio ; PPV- Positive predictive value ; NPV- Negative predictive value ; EBM- Evidence-based medicine  ; IDK- I do not know ; AR- Answer rate ; SD- Standard deviation ; CI- Confidence interval  ; IQR- Interquartile range 

\section{Competing interests}

The authors: E.A., M.L., D.H., D.B.J., M.T.K., D.E., S.L., Y.D., S.B., J.M. and S.O. are paid employees by Kahun Ltd. All other authors declare no financial or non-financial competing interests.

\newpage
\section{Supplementary}
\subsection{Tables}

\renewcommand{\thetable}{S1}

\begin{table}[H] 
\caption{Questions and possible answers for semantic and numeric types}
\centering
\begin{tabular}{|l|p{4cm}|p{4cm}|}
\hline
\textbf{Question type} & \textbf{Question} & \textbf{Possible Answers} \\
\hline
\multirow{3}{*}{Semantic} & Which biological sex has an increased likelihood of Osteoarthritis within the general population? & - Male \newline - Female \newline - I do not know \\
\cline{2-3}
& Which age group/s is/are the most commonly associated with Crohn's disease? & - 60-90 Years \newline - 20-29 Years \newline - I do not know \\
\cline{2-3}
& What is/are the most common location/s of Patch in patients with Mycosis fungoides? & - Breast part \newline - Buttock structure \newline - Lower trunk \newline - Skin structure of inguinal region \newline - I do not know \\
\hline
\multirow{3}{*}{Numeric} & How does Oral contraception influence the chance of Cerebral venous sinus thrombosis? & - Increases the chance by greater than 2.5 times \newline - Increases the chance between 1.01 and 2.5 times \newline - Decreases the chance between 0.7 and 0.99 times \newline - Decreases the chance by less than 0.7 times \newline - I do not know \\
\cline{2-3}
& What is the positive likelihood ratio of dyspnea at rest in patients with Asthma? & - Greater than 3.7 \newline - Between 1.01 and 3.7 \newline - Between 0.35 and 0.99 \newline - Less than 0.35 \newline - I do not know \\
\cline{2-3}
& Is the association between Factor V deficiency and Cerebral venous thrombosis low, medium or high? & - High (greater than 42\% of the cases) \newline - Medium (between 5\% and 42\% of the cases) \newline - Low (less than 5\% of the cases) \newline - I do not know (only if you do not know what the answer is) \\
\hline
\end{tabular}
\end{table}

\FloatBarrier 

\renewcommand{\thetable}{S2}

\begin{table}[H] 
\caption{Distributions of answers according to question type}
\centering
\begin{tabular}{lp{2cm}p{1.5cm}p{2cm}p{1cm}p{1cm}}
\toprule
\textbf{Question type} & \textbf{Median} & \textbf{First quartile} & \textbf{Third quartile} & \textbf{Mean} & \textbf{SD} \\
\midrule
Semantic        & 77  & 73    & 81    & 77.72  & 6.074 \\
Risk Factor     & 81  & 80    & 84    & 82.37  & 3.635 \\
Association     & 68  & 66    & 71    & 68.82  & 4.29  \\
Sensitivity     & 54  & 52    & 58    & 55.488 & 4.19  \\
PLR             & 69  & 66    & 73    & 69.535 & 4.854 \\
Incidence       & 66  & 63.5  & 68    & 66.028 & 3.21  \\
Prevalence      & 58  & 56    & 60    & 58.437 & 3.60  \\
Associated Risk & 72  & 70.5  & 74    & 72.52  & 2.97  \\
Relative Risk   & 71  & 69    & 72.5  & 71.42  & 3.47  \\
NLR             & 69  & 66    & 73    & 69.827 & 5.23  \\
Specificity     & 60  & 55    & 64    & 59.67  & 5.88  \\
PPV             & 62  & 59    & 66.25 & 63.51  & 5.16  \\
NPV             & 62  & 58    & 64    & 62.145 & 5.08  \\
\bottomrule
\end{tabular}
\end{table}
\FloatBarrier 

\renewcommand{\thetable}{S3}

\begin{table}[H] 
\caption{Distributions of answers according to question type}
\centering
\begin{tabular}{|p{1.5cm}|p{1.5cm}|p{1.5cm}|p{1.5cm}|p{1.5cm}|p{1.5cm}|p{1.5cm}|p{1.5cm}|p{1.5cm}}
\toprule
\textbf{} & \textbf{Majority} & \textbf{Claude3 without I do not know} & \textbf{Claude3 with I do not know} & \textbf{GPT4 without I do not know} & \textbf{GPT4 with I do not know} & \textbf{Human without I do not know} & \textbf{Human with I do not know} \\
\hline
\textbf{Claude3 without I do not know} & P = .0163 & NA & P = .1698 & P = .105 & P = .2607 & P = .0001 & P $<$ .0001 \\
\hline
\textbf{Claude3 with I do not know} & P = .003 & P = .1698 & NA & P = .006 & P = .02 & P = .0004 & P $<$.0001 \\
\hline
\textbf{GPT4 without I do not know} & P = .0205 & P = .105 & P = .006 & NA & P = .24 & P $<$ .0001 & P $<$ .0001 \\
\hline
\textbf{GPT4 with I do not know} & P = .002 & P = .2607 & P = .02 & P = .24 & NA & P $<$ .0001 & P $<$ .0001 \\
\bottomrule
\end{tabular}
\end{table}
\FloatBarrier 

\renewcommand{\thetable}{S4}

\begin{longtable}{lp{2cm}p{1.5cm}p{2cm}p{1cm}p{1cm}}
\caption{Proportion comparison according to sub labels} \\
\toprule
Label               & Sub label              & Claude3 Accuracy     & Proportion p-value & GPT4 Accuracy         & 95\% confidence interval \\
\midrule
\endfirsthead
\caption[]{(continued)} \\
\toprule
Label               & Sub label              & Claude3 Accuracy     & Proportion p-value & GPT4 Accuracy         & 95\% confidence interval \\
\midrule
\endhead
\midrule
\endfoot
\bottomrule
\endlastfoot
Answer distribution & 1st Answer             & 47.05\% (2060/ 4378) & \textless{}.0001   & 63.97\% (4438/ 6938)  & {[}-0.19 - 0.15{]}       \\
Answer distribution & 2nd Answer             & 80.96\% (4546/ 5615) & \textless{}.0001   & 47.86\% (4141/ 8653)  & {[}0.32 - 0.35{]}        \\
Answer distribution & 3rd Answer             & 36.85\% (667/ 1810)  & \textless{}.0001   & 64.36\% (1929/ 2997)  & {[}-0.30 - 0.25{]}       \\
Medical discipline  & Cardiovascular System  & 60.2\% (717/ 1191)   & 0.97               & 60.35\% (1076/ 1783)  & {[}-0.04 - 0.04{]}       \\
Medical discipline  & Digestive System       & 60.33\% (844/ 1399)  & 0.27               & 58.43\% (1306/ 2235)  & {[}-0.01 - 0.05{]}       \\
Medical discipline  & Endocrine System       & 59.36\% (298/ 502)   & 0.96               & 59.03\% (428/ 725)    & {[}-0.05 - 0.06{]}       \\
Medical discipline  & Infectious Disease     & 61.7\% (862/ 1397)   & 0.03               & 57.89\% (1207/ 2085)  & {[}0.004 - 0.07{]}       \\
Medical discipline  & Neoplastic Disease     & 68.7\% (676/ 984)    & \textless{}.0001   & 53.5\% (704/ 1316)    & {[}0.11 - 0.19{]}        \\
Medical discipline  & Nervous System         & 60.43\% (681/ 1127)  & 0.01               & 55.54\% (938/ 1689)   & {[}0.01 - 0.09{]}        \\
Medical discipline  & Respiratory System     & 59.17\% (484/ 818)   & 0.21               & 56.29\%  (720/ 1279)  & {[}-0.02 - 0.07{]}       \\
Medical discipline  & Genitourinary System   & 57.78\% (464/ 803)   & 0.11               & 54.1\% (679/ 1255)    & {[}-0.008 - 0.08{]}      \\
Medical subject     & Disorders              & 60.06\% (3181/ 5296) & 0.0004             & 56.88\% (4232/ 7440)  & {[}0.01 - 0.05{]}        \\
Medical subject     & Imaging and Procedures & 64.39\% (463/ 719)   & 0.07               & 60.43\% (1017/ 1683)  & {[}-0.003 - 0.08{]}      \\
Medical subject     & Lab tests              & 64.08\% (619/ 966)   & \textless{}.0001   & 53.45\% (1008/ 1886)  & {[}0.07 - 0.14{]}        \\
Medical subject     & Symptom and Signs      & 63.08\% (2454/ 3890) & \textless{}.0001   & 56.05\% (3438/ 6134)  & {[}0.05 - 0.09{]}        \\
Prevalence          & High                   & 60.46\% (3410/ 5640) & 0.04               & 58.73\% (4842/ 8245)  & {[}0.0006 - 0.03{]}      \\
Prevalence          & Low                    & 64.37\% (598/ 929)   & \textless{}.0001   & 53.21\% (871/ 1637)   & {[}0.07 - 0.15{]}        \\
Prevalence          & Med                    & 63.83\% (2125/ 3329) & \textless{}.0001   & 56.2\% (2919/ 5194)   & {[}0.05 - 0.1{]}         \\
QA length           & Long                   & 58.25\% (3378/ 5799) & \textless{}.0001   & 54.54\% (4418/ 8100)  & {[}0.02 - 0.05{]}        \\
QA length           & Medium                 & 62.74\% (4098/ 6532) & \textless{}.0001   & 57.53\% (6169/ 10723) & {[}0.04 - 0.07{]}        \\
QA length           & Short                  & 66.13\% (1107/ 1674) & 0.0004             & 60.66\% (1451/ 2392)  & {[}0.02 - 0.08{]}        \\
QA Type             & Association            & 55.04\% (906/ 1646)  & 0.0007             & 49.5\% (1146/ 2315)   & {[}0.02 - 0.09{]}        \\
QA Type             & Incidence              & 54.88\% (720/ 1312)  & 0.62               & 53.86\% (802/ 1489)   & {[}-0.03 - 0.05{]}       \\
QA Type             & Risk Factor            & 59.28\% (1268/ 2139) & 0.5                & 58.25\% (1391/ 2388)  & {[}-0.02 - 0.04{]}       \\
QA Type             & Sensitivity            & 63.99\% (5510/ 8611) & \textless{}.0001   & 58.11\% (8394/ 14446) & {[}0.05 - 0.07{]}       
\end{longtable}

\newpage

\subsection{Figures}

\renewcommand{\thefigure}{S1}

\begin{figure}[H] 
    \centering
    \includegraphics[width=1\textwidth]{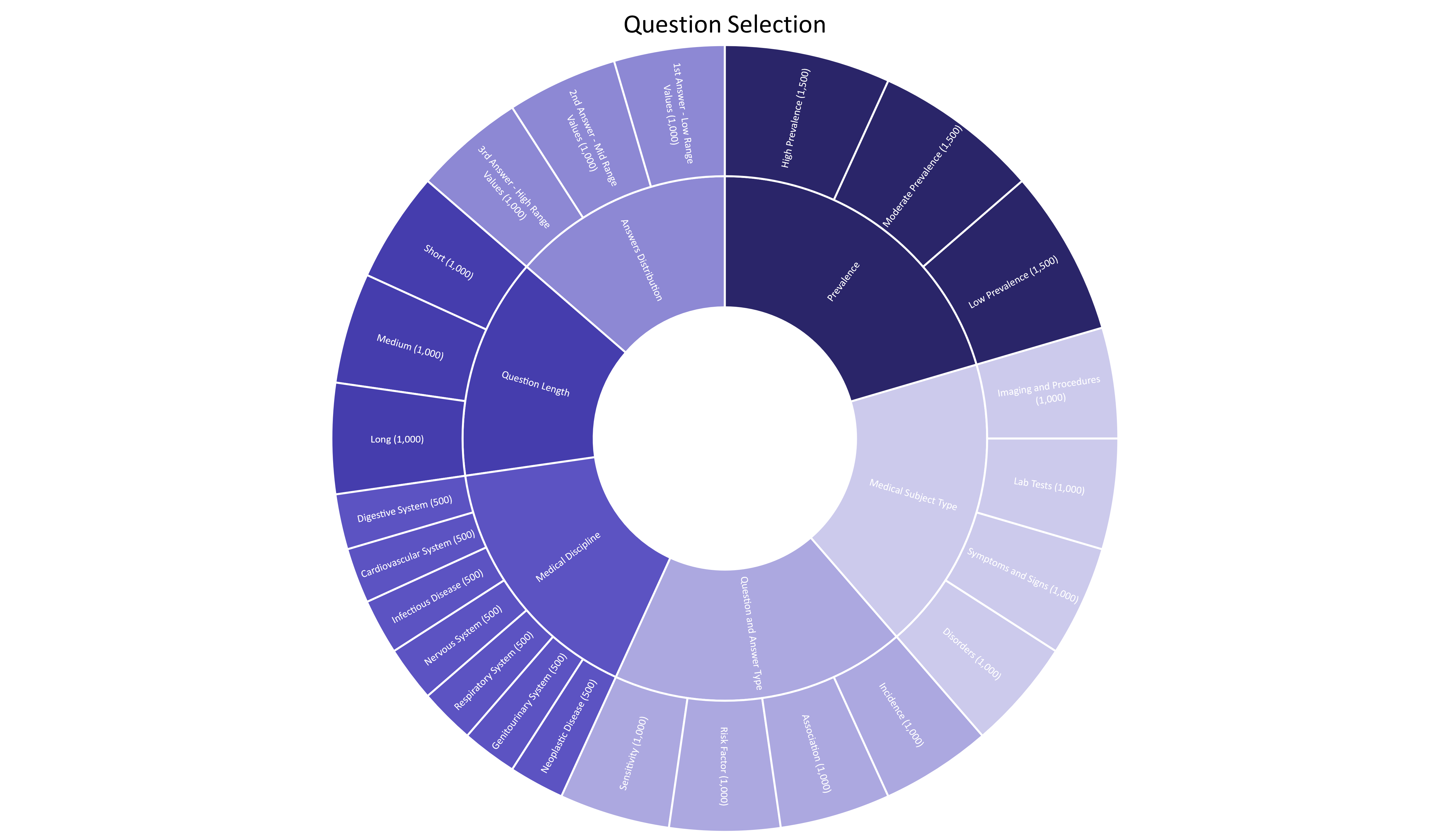}
    \caption{Numeric QAs benchmark subanalysis according to medical and non-medical labels and sub-labels.}
    \label{fig:study_flowchart}
\end{figure}
\FloatBarrier 

\renewcommand{\thefigure}{S2}

\begin{figure}[H] 
    \centering
    \includegraphics[width=1\textwidth]{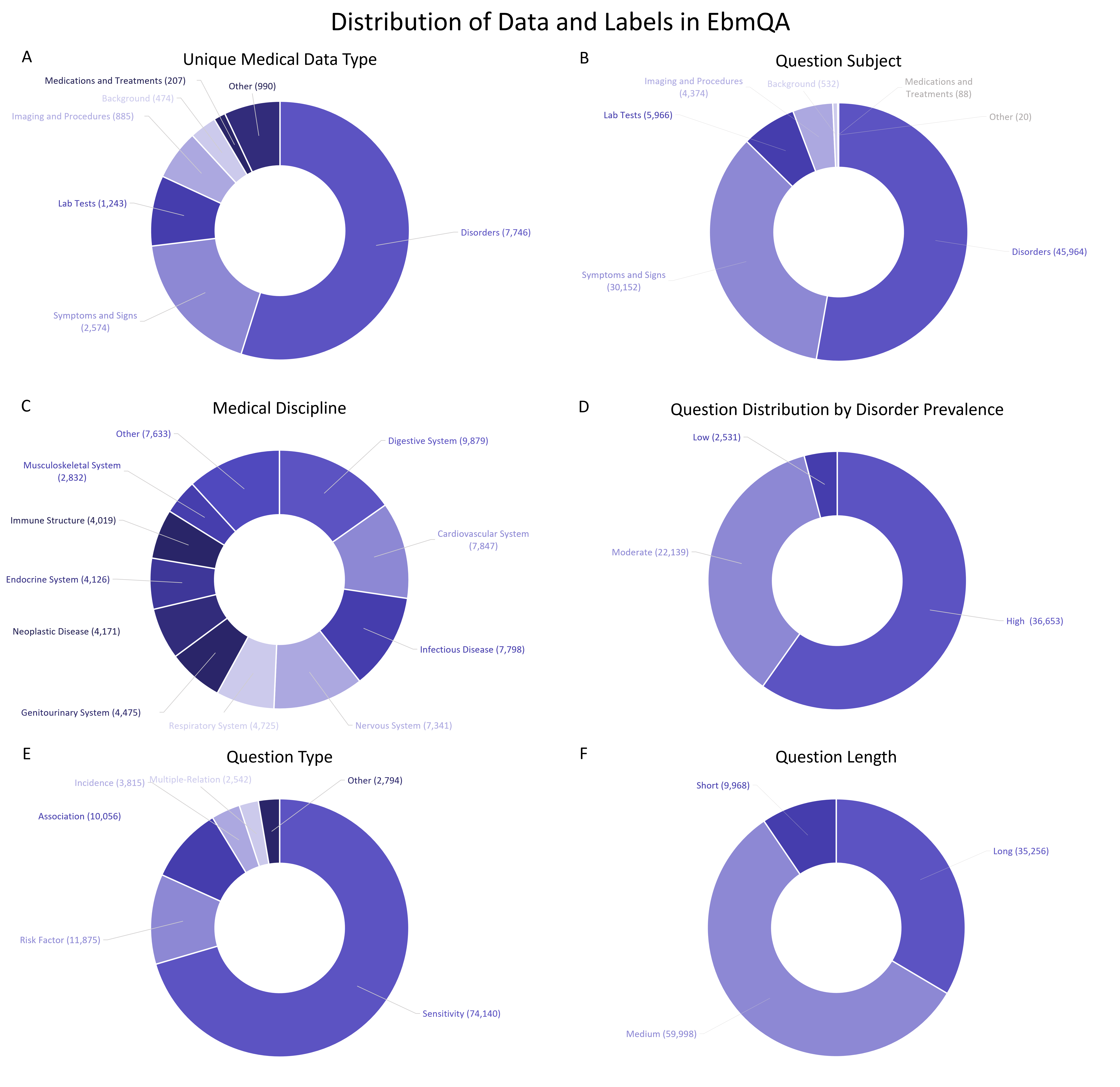}
    \caption{Distribution of the data and labels in the EBMQA: (A) Unique Medical Data Type, (B) Question Subject, (C) Medical Discipline, (D) Disorders Prevalence, (E) Question Type, (F) Question length.}
    \label{fig:study_flowchart}
\end{figure}
\FloatBarrier 

\renewcommand{\thefigure}{S3}
\begin{figure}[H] 
    \centering
    \includegraphics[width=1\textwidth]{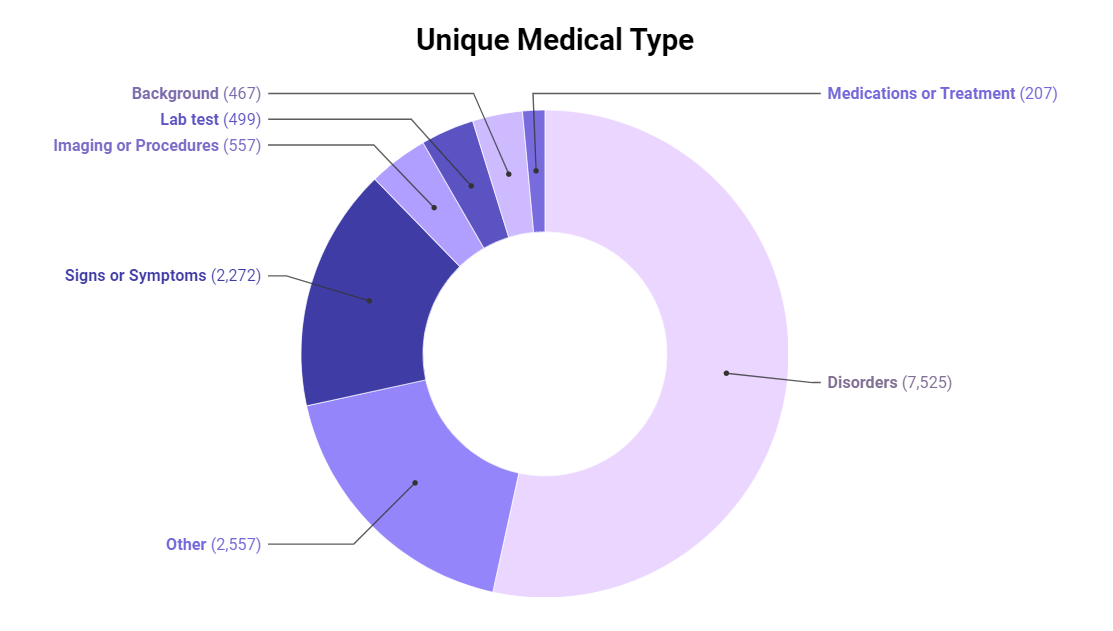}
    \caption{Distribution of the correct answer with mid values ranging from 0-1 of the QAs in the ebmQA by the overall median value of each QA type and the corresponding MAD: 0 $\leq$ mid value $<$ overall median - MAD (Short), overall median - MAD $\leq$ mid value $\leq$ overall median + MAD, (Medium), overall median + MAD $\leq$ mid value $<$ 1 (Long).}
    \label{fig:study_flowchart}
\end{figure}
\FloatBarrier 

\renewcommand{\thefigure}{S4}
\begin{figure}[H] 
    \centering
    \includegraphics[width=1\textwidth]{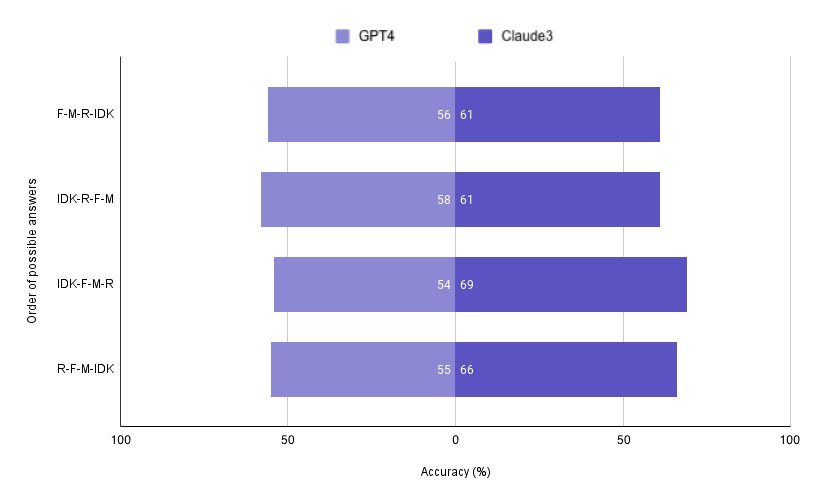}
    \caption{Sensitivity analysis of four prompts with the 'I do not know' option was assessed according to their accuracy. Each row represents a different order of the possible answers. The order of the possible answers in the prompt is based on the sequence of letters/symbols, separated by hyphens, from left to right. Each letter/symbol represents a frequency range determined by the relevant overall median and the median absolute deviation (MAD): frequency range $\geq$ overall median + MAD (F), overall median - MAD $\leq$ frequency range $\leq$ overall median + MAD (M), frequency range $\leq$ overall median - MAD (R), and I do not know (IDK).}
    \label{fig:study_flowchart}
\end{figure}
\FloatBarrier 

\renewcommand{\thefigure}{S5}
\begin{figure}[H] 
    \centering
    \includegraphics[width=1\textwidth]{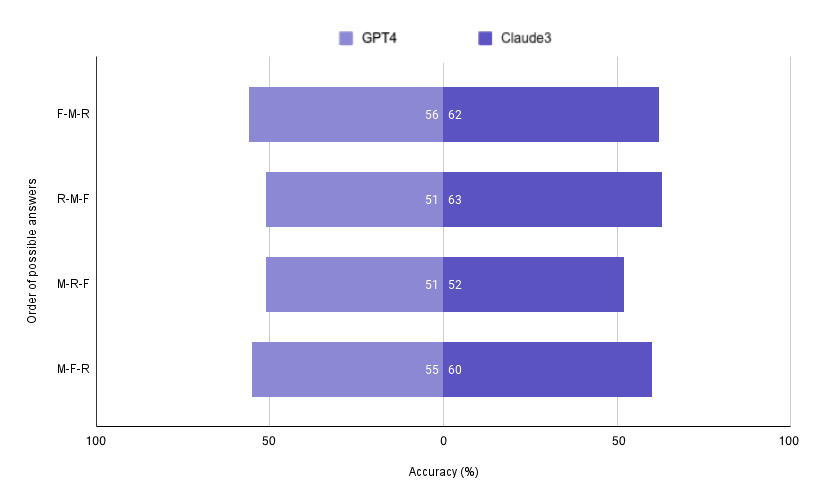}
    \caption{Sensitivity analysis of four prompts without the 'I do not know' option was assessed according to their accuracy. Each row represents a different order of the possible answers. The order of the possible answers in the prompt is based on the sequence of letters/symbols, separated by hyphens, from left to right. Each letter/symbol represents a frequency range determined by the relevant overall median and the median absolute deviation (MAD): frequency range $\geq$ overall median + MAD (F), overall median - MAD $\leq$ frequency range $\leq$ overall median + MAD (M), frequency range $\leq$ overall median - MAD (R).}
    \label{fig:study_flowchart}
\end{figure}
\FloatBarrier 

\newpage
\section{Appendix}

\subsection{EBMQA}
\subsubsection{Template creation}

Kahun's knowledge graph contains nodes, connected through edges. Each\\ evidence-based knowledge (from an article) is represented with two nodes and one edge. For example, node1 contains data regarding a source (for example, a disorder) and its background (a population). Node1 is connected, through edge1, to node2, which contains data regarding a target (for example, a symptom). Edge1 contains data such as the type of connection between node1 and node2, the numerical value of the connection and additional information (for example, a prevalence of $60\%$). For each connection in the graph, a specific template with placeholders for the data from node1 and node2 was designed. As shown in Figure 1, for the "prevalence" type of connection, the template was: "What is the prevalence of target in background with source?". Using this method and the structure of the graph, we were able to create the EBMQA. 

\subsubsection{Numerical data and possible answers}
Since each QA is based on data stored in Kahun's knowledge graph, all the possible answers (correct or wrong) in the semantic QA and the correct answer in the numeric QA are backed with numerical data and citations. Therefore, “wrong” answers in the semantic QA represent connections that exist in the knowledge graph.

The data behind each possible answers (both correct and wrong) in the semantic QA and behind each correct answer in the numeric QA includes the following numerical values:
\begin{enumerate}[label=\textbullet]
    \item Minimum value - a lower estimation of the connection between the source and the target given the concept.
    \item Maximum value - a higher estimation of the connection between the source and the target given the concept.
    \item Mid value - the Mean of the "Minimum value " and the "Maximum value".
\end{enumerate}

In a numeric QA we used the median of all the mid values (overall median) which are related to the specific numeric QA type. We calculated the median absolute deviation (MAD) to form ranges of answers as explained below. The correct answer is a range that includes the Mid value. 
For questions with mid values ranging from 0-1 (such as Prevalence, Association, etc.), four possible answers were defined by a single MAD from the overall median: from 0 to median - MAD (1st answer- low range values), from median - MAD to median + MAD, (2nd answer- mid range values), greater than median + MAD (3rd answer- high range values) and an IDK answer.  
For QA types with Mid values that can be greater than 1 (such as Relative Risk, Odds Ratio, etc.), a value smaller than 1 indicates that the risk of the outcome is decreased by the exposure and a value greater than 1 indicates that the risk of the outcome is increased by the exposure. Therefore, we separated decreasing and increasing values and evaluated the median of values between 0-1 and the median of values greater than 1. As a result, those QA types have five possible answers: from 0 to the decreasing median, from the decreasing median to 1, from 1 to the increasing median, greater than the increasing median and an IDK answer.

On the contrary, a semantic QA can have one or more correct answers out of the possible answers: 
-	A question will have one correct answer if the highest mean value of all possible answers is at least 10\%  higher than the following highest mean value. 
-	A question will have more than one answer if there are several answers with mean values that have less than a 10\%  difference between them and the highest mean value.
-	If all the possible answers have less than a 10\%  difference between them, all answers are considered correct.

\subsubsection {Labeling}

Each QA was classified according to the following medical data labels:
\begin{enumerate}
    \item Medical type - each entity (in the question) was classified according to groups of clinical aspects as suggested by Snomed CT (a medical clinical terminology database used by the U.S. Federal Government) [https://www.  nlm.nih.gov/healthit/snomedct/index.html ] or by Kahun’s medical team. For example: Disorders, Symptoms or Signs, Lab tests etc.
    \item Medical subject type - the label of the “Medical type” of the subject in the question, usually the target’s “Medical type”. Relevant to all types of QAs except form “Association”.
    \item Medical Discipline - entities were classified according to groups of medical topics as suggested by Snomed CT [https://www.nlm.nih.gov/healthit/  snomedct/index.html ] or by Kahun’s medical team. For example: Respiratory system , Neurological system, Cardiovascular system etc.
    \item Unique Medical Discipline - the name of the unique Medical Discipline in the question. This label is relevant only if there is one type of Medical Discipline in the question.
    \item Prevalence - the prevalence of the disorder in the QA as dated in Kahun’s knowledge graph. This label is relevant only if the question includes one disorder, as determined by having a single “Disorder” Medical type label. Inspired by the Head-Torso-Tail approche [Ref], we developed a classification method for disease prevalence. QAs were categorized into widespread, common, and rare disorders based on the overall median prevalence of QAs that have only one disorder. Widespread disorders have a prevalence higher than the median + MAD, rare disorders have a prevalence lower than the median - MAD, and common disorders fall in between these thresholds.
\end{enumerate}
Additionally, each QA was classified according to two main metadata labels: 
\begin{enumerate}
    \item QA type - based on the data from Kahun’s knowledge graph.
    \item Question length - QAs were categorized into long, medium, and short length based on the overall median question length. A long question length is longer than the median + MAD value, a short question length is shorter than the median - MAD, and medium question length falls in between these thresholds.
    \item Answers distribution- Relevant only for QAs with mid values ranging from 0-1 and including three answers: the 1st answer- low range values, the 2nd answer- mid range values and the 3rd answer- high range values.
\end{enumerate}
   
\subsection{Benchmark Analysis}
\subsubsection{QA selection}

Detailed descriptions of the QAs in the benchmark (Figure S1):
\begin{enumerate}
    \item Medical subject type (medical): four medical subject types were selected, each with 1,000 QAs per type: “Disorders”, “Signs or Symptoms”, “Imaging or Procedures” and “Lab Test”.
    \item Medical Discipline (medical): seven medical discipline types were selected, each with 500 QAs per type: Digestive system, Cardiovascular system, Infectious disease, Nervous system, Respiratory system, Genitourinary system and Neoplastic disease.
    \item Prevalence (medical): An equal number of 1,500 QAs were selected from each of the three “Prevalence” categories.
    \item QA types (non-medical): four QA types were selected, each with 1,000 QAs per type: “Sensitivity”,”Risk Factor”, “Association” and “Incidence”.
    \item Question length (non-medical): An equal number of 1,000 QAs were selected from each of the three “Question length” categories.
    \item Answers distribution (non-medical): An equal number of 1,000 QAs were selected from each of the three “Answers distribution” categories.
\end{enumerate}

\subsubsection {LLMs prompting}

The prompt for each QA included the question itself, a specific text asking to choose the answer only from the provided possible answers while not adding additional text. Moreover, an IDK option was added to the possible answers.
An example for the semantic prompt:
Among the possible answers, What is/are the most common subtype/s of Dementia? The possible answers are: ' Alzheimer's disease  ' , ' Multi-infarct dementia  ' , ' Diffuse Lewy body disease  ' , ' Frontotemporal dementia  ' . You must base your response exclusively on the possible answers provided. No other words or answers are allowed. You can choose a single answer or multiple answers from the answers provided. If you do not know the answer to the question, respond with ‘I do not know’.
An example for the numeric prompt:
What is the prevalence of dysuria in female patients? Choose the correct answer from the following options, without adding further text: (1) Greater than 54\% , (2) Between 5\%  and 54\% , (3) Less than 5\% , (4) I do not know (only if you do not know what the answer is).

\end{document}